\pdfoutput=1
\documentclass{article}

\usepackage[preprint]{neurips_2023}

\usepackage[utf8]{inputenc} 
\usepackage[T1]{fontenc}    
\usepackage{hyperref}       
\usepackage{url}            
\usepackage{booktabs}       
\usepackage{amsfonts}       
\usepackage{nicefrac}       
\usepackage{microtype}      
\usepackage[table,xcdraw]{xcolor}         

\usepackage{tabularx,booktabs}
\newcolumntype{Y}{>{\centering\arraybackslash}X}

\usepackage{subcaption}
\usepackage{wrapfig}
\usepackage{multirow}
\usepackage{makecell}
\usepackage{xspace}
\usepackage{enumitem}
\usepackage{amsmath}
\usepackage{listings}
\usepackage{algorithm}
\usepackage{algorithmic}
\usepackage{amsfonts,amssymb}
\usepackage{mathrsfs}
\usepackage{subcaption}
\usepackage{natbib}
\setcitestyle{numbers,square}
\usepackage{graphicx}       
\usepackage{cleveref}
\usepackage{hyperref}

\newcommand{\hhide}[1]{}
\newcommand{\hide}[1]{}

\newcommand{\vpara}[1]{\vspace{0.07in}\noindent\textbf{#1}\xspace} %

\title{ChatGLM: A Family of Large Language Models \\from GLM-130B to GLM-4 All Tools}

 \author{
Team GLM\\ \\
\textsuperscript{1}Zhipu AI \quad
\textsuperscript{2}Tsinghua University \quad 
\\ \\
{\includegraphics[height=3.5ex]{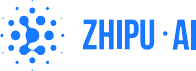}}
}

\begin{document}

\maketitle
\renewcommand{\thefootnote}{\fnsymbol{footnote}}
    \footnotetext[1]{Team GLM: Aohan Zeng, Bin Xu, Bowen Wang, Chenhui Zhang, Da Yin, Dan Zhang, Diego ROJAS, Guanyu Feng, Hanlin Zhao, Hanyu Lai, Hao Yu, Hongning Wang, Jiadai Sun, Jiajie Zhang, Jiale Cheng, Jiayi Gui, Jie Tang, Jing Zhang, Jingyu Sun, Juanzi Li, Lei Zhao, Lindong Wu, Lucen Zhong, Mingdao Liu, Minlie Huang, Peng Zhang, Qinkai Zheng, Rui Lu, Shuaiqi Duan, Shudan Zhang, Shulin Cao, Shuxun Yang, Weng Lam Tam, Wenyi Zhao, Xiao Liu, Xiao Xia, Xiaohan Zhang, Xiaotao Gu, Xin Lv, Xinghan Liu, Xinyi Liu, Xinyue Yang, Xixuan Song, Xunkai Zhang, Yifan An, Yifan Xu, Yilin Niu, Yuantao Yang, Yueyan Li, Yushi Bai, Yuxiao Dong, Zehan Qi, Zhaoyu Wang, Zhen Yang, Zhengxiao Du, Zhenyu Hou, Zihan Wang. 
    }
    \footnotetext[2]{Team members are listed alphabetically by first name.}
\renewcommand{\thefootnote}{\arabic{footnote}}

\begin{abstract}

We introduce ChatGLM, an evolving family of large language models that we have been developing over time. 
This report primarily focuses on the GLM-4 language series, which includes GLM-4, GLM-4-Air, and GLM-4-9B. 
They represent our most capable models that are trained with all the insights and lessons gained from the preceding three generations of ChatGLM. 
To date, the GLM-4 models are pre-trained on ten trillions of tokens mostly in Chinese and English, along with a small set of corpus from 24 languages, and aligned primarily for Chinese and English usage. 
The high-quality alignment is achieved via a multi-stage post-training process, which involves supervised fine-tuning and learning from human feedback. 
Evaluations show that GLM-4, 1) closely rivals or outperforms GPT-4 in terms of general metrics such as MMLU, GSM8K, MATH, BBH, GPQA, and HumanEval, 2) gets close to GPT-4-Turbo in instruction following as measured by IFEval, 3) matches GPT-4 Turbo (128K) and Claude 3 for long context tasks, and 4) outperforms GPT-4 in Chinese alignments as measured by AlignBench. 
The GLM-4 All Tools model is further aligned to understand user intent and autonomously decide when and which tool(s) to use---including web browser, Python interpreter, text-to-image model, and user-defined functions---to effectively complete complex tasks. 
In practical applications, it matches and even surpasses GPT-4 All Tools in tasks like accessing online information via web browsing and solving math problems using Python interpreter. 
Over the course, we have open-sourced a series of models, including ChatGLM-6B (three generations), GLM-4-9B (128K, 1M), GLM-4V-9B, WebGLM, and CodeGeeX, attracting over 10 million downloads on Hugging face in the year 2023 alone. 
The open models can be accessed through \url{https://github.com/THUDM} and \url{https://huggingface.co/THUDM}. 

\end{abstract}
\clearpage

\begin{figure}[tb]
    \centering
    \includegraphics[width=\linewidth]{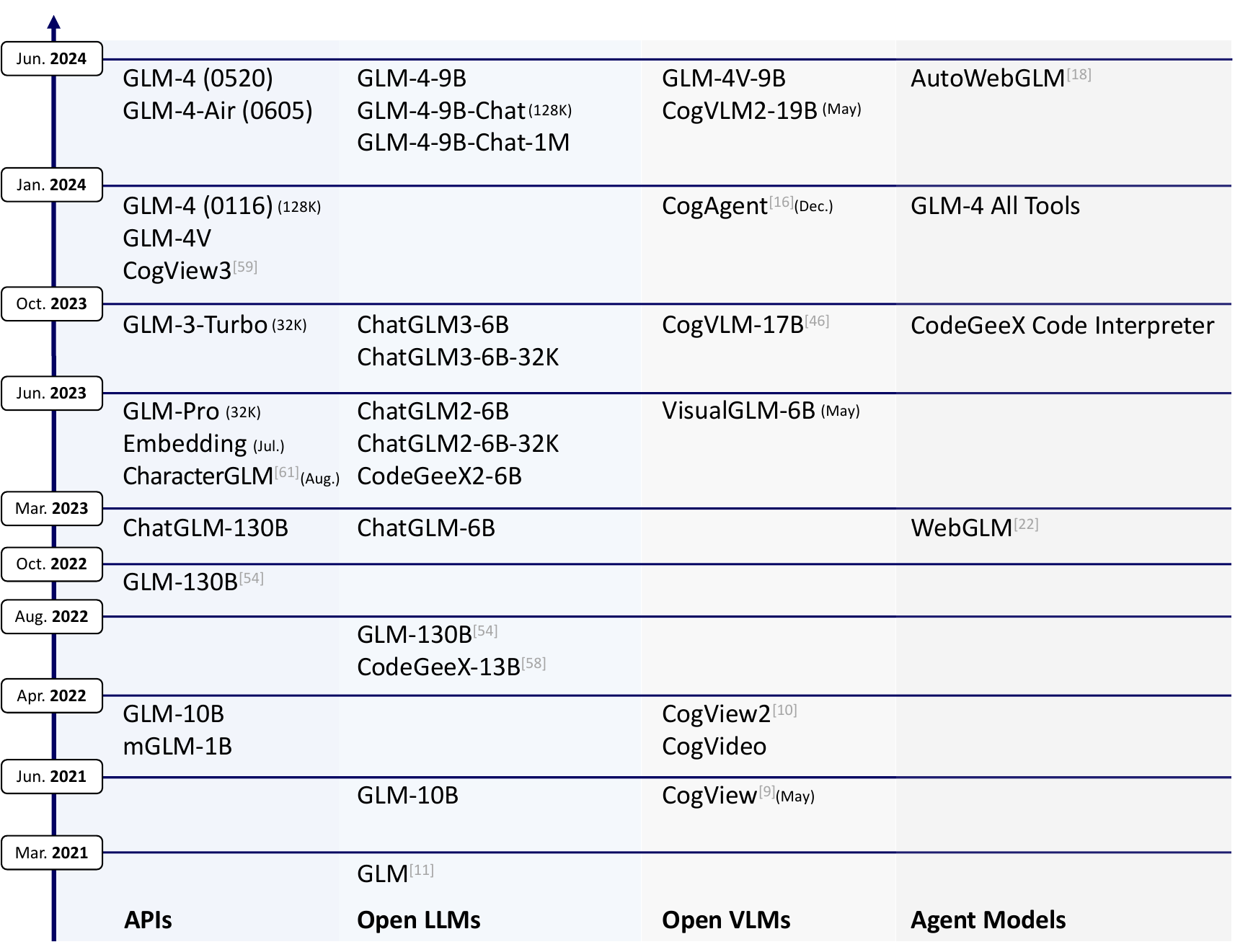}
    \caption{Timeline of the GLM family of language, code, vision, and agent models. The focus of this report is primarily on the language models, i.e., ChatGLM. The APIs are publicly available at \url{https://bigmodel.cn} and open models can be accessed through \url{https://github.com/THUDM}.}
    \label{fig:timeline}
\end{figure}

\hide{
\renewcommand{\thefootnote}{}
    \footnotetext[1]{GLM: \url{https://github.com/THUDM/GLM}}
    \footnotetext[2]{CogView: \url{https://github.com/THUDM/CogView}}
    \footnotetext[3]{CogView2: \url{https://github.com/THUDM/CogView2}}
    \footnotetext[4]{CodeGeex: \url{https://github.com/THUDM/CodeGeeX}}
    \footnotetext[5]{GLM-130B: \url{https://github.com/THUDM/GLM-130B}}
    \footnotetext[6]{ChatGLM-6B: \url{https://github.com/THUDM/ChatGLM-6B}}
    \footnotetext[7]{WebGLM: \url{https://github.com/THUDM/WebGLM}}
    \footnotetext[8]{ChatGLM2-6B: \url{https://github.com/THUDM/ChatGLM2-6B}}
    \footnotetext[9]{CodeGeex2: \url{https://github.com/THUDM/CodeGeeX2}}
    \footnotetext[10]{CharacterGLM-6B: \url{https://github.com/thu-coai/CharacterGLM-6B}}
    \footnotetext[11]{CogVLM \& CogAgent: \url{https://github.com/THUDM/CogVLM}}
    \footnotetext[12]{ChatGLM3-6B: \url{https://github.com/THUDM/ChatGLM2-6B}}
    \footnotetext[13]{CogVLM2: \url{https://github.com/THUDM/CogVLM2}}
    \footnotetext[14]{AutoWebGLM: \url{https://github.com/THUDM/AutoWebGLM}}
    \footnotetext[15]{GLM-4: \url{https://github.com/THUDM/GLM-4}}
\renewcommand{\thefootnote}{\arabic{footnote}}
}

\section{Introduction}


The rapid development of large language models (LLMs) has been phenomenal \cite{zhao2023survey}. Take one of the most successful model series, the OpenAI's GPT models, as an example: the original GPT-3 model released in 2020~\cite{GPT3} marked a significant scale-up from GPT-1's 117 million parameters and GPT-2's 1.5 billion parameters, to 175 billion parameters. 
This scale-up enables the decoder-only transformer-based GPT-3 model with in-context learning and generalized capabilities: according to OpenAI, the GPT-3.5 series improved upon GPT-3 by incorporating instruction tuning, supervised fine tuning (SFT), and/or reinforcement learning from human feedback (RLHF)~\cite{ouyang2022training}. This has now became a standard procedure to create performing LLMs, including the PaLM models \cite{chowdhery2022palm}, the LLaMA models \cite{touvron2023llama}, the Gemini models \cite{geminiteam2023gemini}, and many more.  

In a parallel line to the popularly adopted LLMs development practices, we proposed the General Language Model (GLM) architecture~\cite{du2022glm} featured with the autoregressive blank infilling objective and open-sourced the GLM-10B model in 2021 (See the GLM timeline in Figure \ref{fig:timeline}). 
Starting in late 2021, we began pre-training GLM-130B~\cite{zeng2022glm}. 
The goal was to train a 100B-scale model to match or surpass GPT-3 (davinci) while also verifying the techniques for successfully training models at this scale, along with other contemporary efforts such as OPT-175B~\cite{zhang2022opt} and BLOOM-176B~\cite{scao2022bloom}. 
We completed the 400B-token training and evaluation of GLM-130B in July, and subsequently released the model and pre-training details~\cite{zeng2022glm} in August 2022. 
According to HELM in November 2022, GLM-130B matches GPT-3 (davinci) across various dimensions~\cite{liang2023holistic}.

Following this, we initiated instruction tuning on GLM-130B. 
Later, ChatGPT further motivated us to align the base models with SFT and RLHF. 
We created and crafted the prompt-response pairs from scratch and performed SFT, while also starting to examine how to effectively apply RLHF. 
On March 14, 2023, the aligned model, ChatGLM-130B, went live on \url{https://chatglm.cn}. 
In addition, a smaller version, ChatGLM-6B~\cite{github:chatglm-6b}, was open-sourced on the same day, attracting significantly more attention than anticipated. 
It was designed to have 6.2 billion parameters for 1) facilitating fast iteration of pre-and post-training techniques as well as data selection, and 2) enabling local deployment on consumer-grade graphics cards using INT4 quantization. 
Since then, we have been rapidly exploring and refining our pre-training and alignment techniques, leading to the second and third generations of ChatGLM series every other three months, both of which were pre-trained entirely from the beginning.

ChatGLM-6B was pre-trained on approximately one trillion tokens of Chinese and English corpus with a context length of 2,048 (2K), supplemented mostly by SFT.
Released in June, ChatGLM2-6B was pre-trained and aligned with more high-quality data, leading to substantial improvements over its predecessor, including a 23\% improvement on MMLU, 571\% on GSM8K, and 60\% on BBH. 
By adopting the FlashAttention technique~\cite{dao2022flashattention}, its context length was extended to 32K. 
Additionally, the integration of Multi-Query Attention~\cite{shazeer2019fast} contributed to a 42\% increase in inference speed. 
Taking this further, our 2nd generation code model CodeGeeX2-6B was developed by pre-training on an additional 600 billion code tokens. 
It demonstrated Pass@1 improvements over the initial generation, CodeGeeX-13B~\cite{zheng2023codegeex}, with increases of 57\% in Python, 71\% in C++, 54\% in Java, 83\% in JavaScript, and 56\% in Go  as measured by HumanEval-X. 
When adapting to Character-based Dialogues, CharacterGLM~\cite{zhou2023characterglm} allows effective and safe character customization on LLMs.
By further adapting more diverse training datasets, more sufficient training steps, and more optimized training strategies, ChatGLM3-6B topped 42 benchmarks across semantics, mathematics, reasoning, code, and knowledge. 
Starting from this generation, ChatGLM also supports function call and code interpreter, as well as complex agent tasks~\cite{liu2023webglm,zeng2023agenttuning,lai2024autowebglm}. 
In the course of these developments, we also developed models with 1.5B, 3B, 12B, 32B, 66B, and 130B parameters, allowing us to validate our observations and establish our own scaling laws.

With all the lessons learned and experiences accumulated, we kicked off the training of GLM-4. 
The first cutoff checkpoint then underwent a multi-stage post-training process (e.g., SFT, RLHF, safety alignment) with a focus on the Chinese and English language for now. 
Subsequently, it was developed into two distinct versions: GLM-4 and GLM-4 All Tools, both supporting a 128K context length. 
Since Janurary 16, 2024, GLM-4 (0116) has been made available through the GLM-4 API at \url{https://bigmodel.cn}, and GLM-4 All Tools is accessible via the website \url{https://chatglm.cn} and mobile applications that support the creation of one's own agent---GLMs. 
The latest models are GLM-4 (0520) and GLM-4-Air (0605) with an upgrade on both pre-training and alignment. 
GLM-4-Air achieves comparable performance to GLM-4 (0116) with lower latency and inference cost. 
Evaluations of GLM-4 were performed on a variety of language benchmarks. 
These evaluations assess GLM-4's general abilities in English, instruction following in both English and Chinese, and alignment, long-context, and agent capacities in Chinese.  

\begin{figure}[tb]
    \centering
    \includegraphics[width=\linewidth]{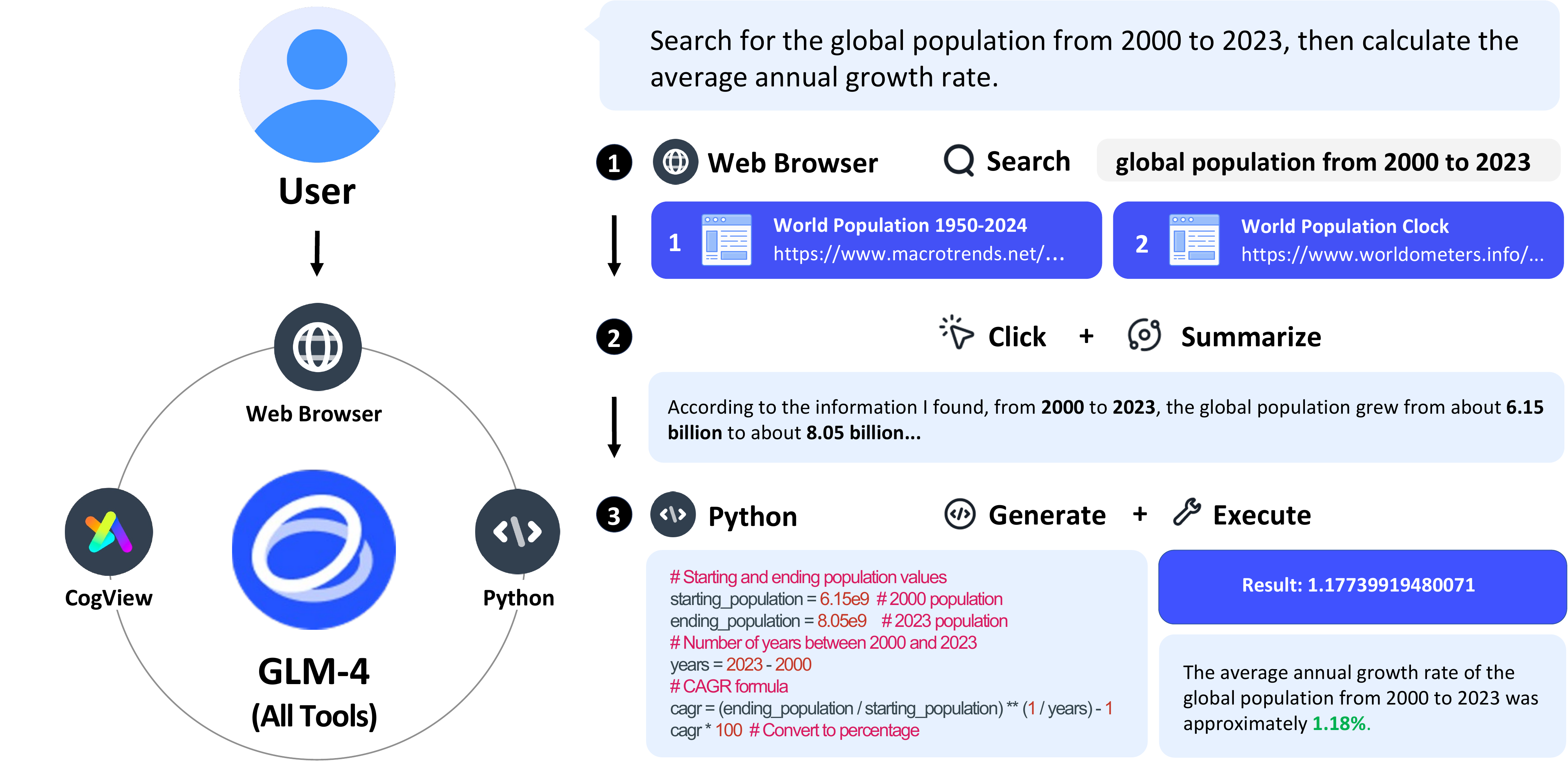}
    \caption{An Illustrative Example of GLM-4 All Tools.
    }
    \label{fig:alltools-example}
\end{figure}

First, on the most commonly-used English academic benchmarks---MMLU, GSM8K, MATH, BBH, GPQA, and HumanEval, GLM-4 0520 achieves performance closely comparable to that of GPT-4 0613~\cite{openai2023gpt} and Gemini 1.5 Pro~\cite{geminiteam2023gemini}. 
For example, it scores 83.3 vs. 86.4 and 83.7 on MMLU, respectively. 
Second, according to IFEval~\cite{zhou2023instruction}, GLM-4's instruction following capacities on both prompt and instruction levels are approximately as effective as GPT-4-Turbo in both English and Chinese. 
Third, in terms of  Chinese language alignment, GLM-4 outperforms GPT-4 and matches GPT-4-Turbo across eight dimensions in AlignBench~\cite{liu2023alignbench}.   
Finally, for long-context tasks, the GLM-4 (128K) model matches the performance of GPT-4 Turbo and Claude 3 Opus as measured by LongBench-Chat~\cite{bai2024longalign}, i.e., 87.3 vs. 87.2 and 87.7, respectively.

\begin{table}[t]
\centering
\renewcommand{\arraystretch}{1.4}
\caption{Performance of Open ChatGLM-6B, ChatGLM2-6B, ChatGLM3-6B, and GLM-4-9B.}
\resizebox{\textwidth}{!}{
\begin{tabular}{@{}ll|rrrr@{}}
\toprule[1.2pt]
Language                  & Dataset          & ChatGLM-6B & ChatGLM2-6B & ChatGLM3-6B-Base & GLM-4-9B \\
                          &                  & \small{(2023-03-14)} &
\small{(2023-06-25)} & \small{(2023-10-27)} & \small{(2024-06-05)} \\
\midrule
\multirow{6}{*}{English} & GSM8K            & 1.5        & 25.9        & 72.3  & 84.0             \\
                          & MATH             & 3.1        & 6.9         & 25.7  & 30.4           \\
                          & BBH              & 0.0        & 29.2        & 66.1    & 76.3         \\
                          & MMLU             & 25.2       & 45.2        & 61.4 & 74.7             \\
                          & GPQA & - & - & 26.8 & 34.3 \\
                          & HumanEval        & 0.0        & 9.8         & 58.5    & 70.1         \\
                          & BoolQ            & 51.8       & 79.0        & 87.9   & 89.6          \\
                          & CommonSenseQA    & 20.5       & 65.4        & 86.5  & 90.7           \\
                          & HellaSwag        & 30.4       & 57.0        & 79.7  & 82.6           \\
                          & PIQA             & 65.7       & 69.6        & 80.1  & 79.1           \\
                          & DROP             & 3.9        & 25.6        & 70.9   & 77.2          \\ 
                          \midrule
\multirow{3}{*}{Chinese} & C-Eval           & 23.7       & 51.7        & 69.0  &     77.1    \\
                          & CMMLU            & 25.3       & 50.0        & 67.5 & 75.1             \\
                          & GAOKAO-Bench     & 26.8       & 46.4        & 67.3 & 74.5            \\
                          & C3               & 35.1       & 58.6        & 73.9  & 77.2         \\
                          \bottomrule[1.2pt]
\end{tabular}
}
\label{tab:6b-performance}
\end{table}

The GLM-4 All Tools model is specifically aligned to better understand user intent and autonomously select the most appropriate tool(s) for task completion. 
For example, it can access online information via a web browser in a multi-round manner, use Python interpreter to solve math problems, leverage a text-to-image model to generate images, and  call user-defined functions. 
Figure \ref{fig:alltools-example} illustrates an example showing GLM-4 All Tools with a web browser and Python interpreter for addressing the user query of ``Search for the global population from 2000 to 2023, then calculate the average annual growth rate''.  
Our first-hand test shows that it not only matches but often surpasses the capabilities of GPT-4 All Tools for common tasks.

Following our three generations of open ChatGLM-6B models, we also openly released the GLM-4-9B (128K and 1M context length) model. 
GLM-4-9B is pre-trained on approximately ten trillion tokens of multilingual corpus with a context length of 8192 (8K) and post-trained with the same pipeline and data used for GLM-4 (0520). 
With less training compute, it outperforms Llama-3-8B~\cite{llama3} and  supports all the functionality of All Tools in GLM-4. 
We also provide an experimental model GLM-4-9B-Chat-1M with 1 million (1M) context length (about 2 million Chinese characters). 
\Cref{tab:6b-performance} shows the performance of the three generations of ChatGLM-6B models and GLM-4-9B, illustrating the progressive improvements of ChatGLM over time.

\begin{figure}[!ht]
    \centering
    \includegraphics[width=.75\linewidth]{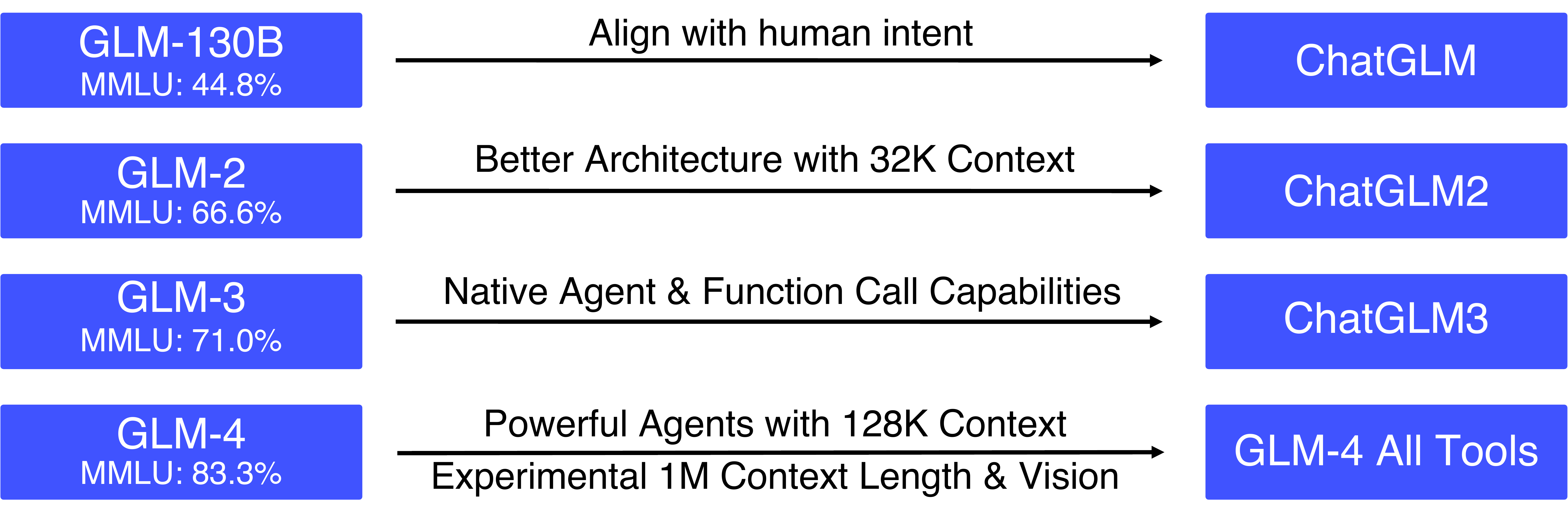}
    \caption{From GLM-130B to ChatGLM to ChatGLM2/3 to GLM-4 All Tools. 
    }
    \label{fig:glm130-glm4at}
\end{figure}

Figure \ref{fig:glm130-glm4at} summarizes the major improvements and features from GLM-130B to GLM-4 All Tools. 
Throughout this journey, we have also contributed to the open development of the code LLMs (CodeGeeX~\cite{zheng2023codegeex}) as well as visual language models for image understanding (CogVLM~\cite{wang2023cogvlm} and CogAgent~\cite{hong2023cogagent}) and text-to-image generation (CogView~\cite{ding2021cogview,ding2022cogview2,zheng2024cogview3}). 
The open models and data can be accessed via \url{https://github.com/THUDM} and \url{https://huggingface.co/THUDM}.

\section{ChatGLM Techniques}

In this section, we introduce both the pre- and post-training techniques we adopted and developed in ChatGLM, including the model architecture, pre-training data, alignment, and All Tools. 
We have detailed technical reports introducing each of the major techniques we used to reach GLM-4.

\vpara{Pre-Training Data.} 
Our pre-training corpus consists of multilingual (mostly English and Chinese) documents from a mixture of different sources, including webpages, Wikipedia, books, code, and research papers. 
The data processing pipeline mainly includes three stages: deduplication, filtering, and tokenization. 
The deduplication stage improves data diversity by removing duplicated or similar documents, with both exact and fuzzy deduplication. 
The filtering stage for webpages improves data quality by removing noisy documents that contain offensive language, placeholder text, source code, etc. 
The tokenization stage converts text into a sequence of tokens for further processing. 
The number of tokens in the pre-training data directly affects model training speed. 
To optimize this aspect, we employ the byte-level byte pair encoding (BPE) algorithm~\cite{sennrich2016neural} to separately learn the Chinese and multilingual tokens and merge them with the tokens of the cl100k\_base tokenizer in tiktoken~\cite{tiktoken} into a unified vocabulary with a size of 150,000. 
In the final training set, we re-weight different sources to increase the importance of high-quality and educational sources like books and Wikipedia. To this end, the pre-training corpus consists of around ten trillion tokens.

Throughout the four generations of ChatGLM development, our findings align with existing studies~\cite{zhou2023lima}: data quality and diversity are crucial for building effective LLMs. Despite the empirical lessons and insights gained, we have to date yet to identify a fundamental principle that could guide the processes of data collection, cleaning, and selection, which might inspire future research directions.

\vpara{Architecture.} 
The GLM family of LLMs is built on Transformer~\cite{vaswani2023attention}. 
In GLM-130B~\cite{zeng2022glm}, we explored various options to stabilize its pre-training by taking into account the hardware constraints we faced at the time. 
Specifically, GLM-130B leveraged DeepNorm~\cite{wang2022deepnet} as the layer normalization strategy and used Rotary Positional Encoding (RoPE)~\cite{su2021roformer} as well as the Gated Linear Unit~\cite{shazeer2020glu} with GeLU~\cite{hendrycks2016gaussian} activation function in FFNs. 
Throughout our exploration, we have investigated different strategies to enhance model performance and inference efficiency.  
The recent GLM-4 model adopts the following architecture design choices.

\begin{itemize}[leftmargin=*,itemsep=0pt,parsep=0.2em,topsep=0.0em,partopsep=0.0em]
\item \textbf{No Bias Except QKV}: 
To increase training speed, we removed all bias terms with the exception of the biases in Query, Key, and Value (QKV) matrices of the attention layers. 
In doing so, we observed a slight improvement in length extrapolation.

\item \textbf{RMSNorm and SwiGLU}: 
We adopted RMSNorm and SwiGLU to replace LayerNorm and ReLU, respectively. 
These two strategies brought better model performance. 

\item \textbf{Rotary positional embeddings (RoPE)}: We extended the RoPE to a two-dimensional form to accommodate the 2D positional encoding in GLM.

\item \textbf{Group Query Attention (GQA)}: 
We replaced Multi-Head Attention (MHA) with Group Query Attention (GQA) to cut down on the KV cache size during inference. 
Given GQA uses fewer parameters than MHA, we increased the FFN parameter count to maintain the same model size, i.e., setting $d_{\mathrm{ffn}}$ to 10/3 of the hidden size.
\end{itemize}

The context length of our models was extended from 2K (ChatGLM), to 32K (ChatGLM2 and ChatGLM3), and to 128K and 1M (GLM-4). 
These expansions were achieved not only through context extension---position encoding extension ~\cite{press2022train,chen2023extending} and continual training ~\cite{xiong2023effective} on long text---but also long context alignment, enabling GLM-4 to effectively handle very long contexts (Cf ~\cite{bai2024longalign} for technical details).

\vpara{Alignment.}
Pre-training builds the foundation of LLMs while post-training~\cite{ouyang2022training} further refines these models to align with human preferences, such as understanding human intents, following instructions, and facilitating multi-turn dialogues. 
For GLM-4, the alignment is mostly achieved with supervised fine-tuning (SFT) and reinforcement learning from human feedback (RLHF)~\cite{hou2024chatglmrlhf}. 
In SFT, we find that authentic human prompts and interactions instead of template-based or model-generated responses are vital to the alignment quality. 
While SFT largely aligns the base models with human preferences, RLHF can further help mitigate  issues of response rejection, safety, mixture of bilingual tokens generated, and multi-turn coherence among others.

For the first generation of our models (ChatGLM-6B and ChatGLM-130B), the prompt-response pairs were mostly annotated by the model developers. 
For later models, the alignment data is a combination of in-house annotation and proprietary data acquired from third parties, subject to strict quality control measures. 
Similar to existing practices \cite{touvron2023llama2}, annotators are instructed to score model responses from several dimensions, including safety, factuality, relevance, helpfulness, and human preferences.

\vpara{ChatGLM Techniques.}
Throughout the development of ChatGLM, we have introduced and will publish techniques that are used to enhance its performance. 
\begin{itemize}[leftmargin=*,itemsep=0pt,parsep=0.2em,topsep=0.0em,partopsep=0.0em]

\item \textbf{Emergent Abilities of LLMs~\cite{du2024understanding}}: 
We examined the relationship between pre-training loss and performance on downstream tasks and found that with the same pre-training loss, LLMs of different model sizes and training tokens generate the same downtream performance. We also found that on some tasks (such as MMLU and GSM8K), the performance improves beyond random chance only when the pre-training loss falls below a certain threshold.
We thus redefine emergent abilities as those exhibited by models with lower pre-training losses~\cite{du2024understanding}.

\item \textbf{LongAlign~\cite{bai2024longalign}}: 
To extend LLMs' context window size, we proposed LongAlign---a comprehensive recipe for long context alignment. 
It enables GLM-4 to process long context texts (up to 128K tokens) with performance comparable to that of Claude 2 and GPT-4 Turbo (1106). 

\item \textbf{ChatGLM-Math~\cite{xu2024chatglmmath}}: 
To improve math problem solving in LLMs, we introduced ChatGLM-Math that leverages self-critique rather than external models or manual annotations for data selection. 

\item \textbf{ChatGLM-RLHF~\cite{hou2024chatglmrlhf}}: 
To align LLMs with human feedback, we introduced ChatGLM-RLHF---our practices of applying PPO and DPO into LLMs. 

\item \textbf{Self-Contrast~\cite{liu2024selfcontrast}}: 
To avoid the need for expensive human preference feedback data, we developed a feedback-free alignment strategy Self-Contrast. 
It utilizes the target LLM to self-generate massive negative samples for its RLHF alignment. 

\item \textbf{AgentTuning~\cite{zeng2023agenttuning}}: 
To improve LLMs' agent capabilities, we developed the AgentTurning framework with the  AgentInstruct instruction-tuning dataset that includes high-quality interaction trajectories between agents and environment. 

\item \textbf{APAR~\cite{Liu2024APAR}}: 
To improve the inference speed of LLMs for responses with hierarchical structures, we presented an auto-parallel auto-regressive (APAR) generation approach. 
It leverages instruct tuning to train LLMs to plan their (parallel) generation process and execute APAR generation. 

\item \textbf{Benchmarks}: 
We also developed several open LLM benchmarks, including 
AgentBench~\cite{liu2023agentbench} for evaluating LLMs as agents, 
LongBench~\cite{bai2023longbench} for evaluating the long context handling performance of LLMs, 
AlignBench~\cite{bai2024longalign} to measure the alignment quality of ChatGLM with Chinese language content, 
HumanEval-X~\cite{zheng2023codegeex} to evaluate HumanEval~\cite{Humaneval:abs-2107-03374} problems in programming languages beyond Python, 
as well as 
NaturalCodeBench (NCB) to measure models' capacities to solve practical programming tasks. 

\end{itemize}

\begin{figure}[tb]
    \centering
    \includegraphics[width=\linewidth]{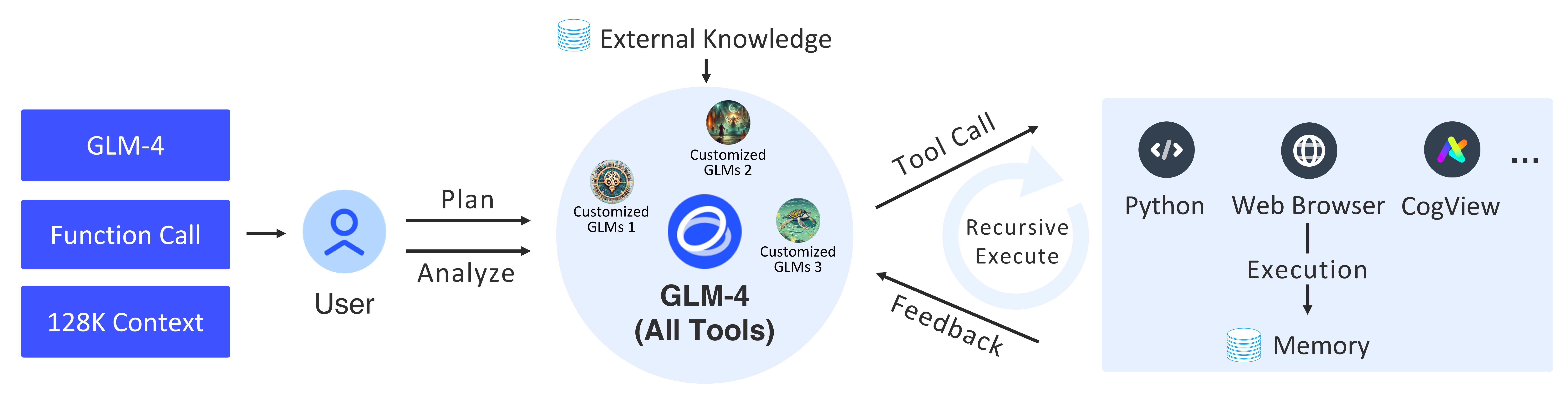}
    \caption{The overall pipeline of GLM-4 All Tools and customized GLMs (agents). 
    }
    \label{fig:alltools-arch}
\end{figure}

\vpara{GLM-4 All Tools.}
The latest ChatGLM models are GLM-4 and GLM-4 All Tools, both of which were trained and aligned by using the techniques above. 
GLM-4 All Tools is a model version further aligned to support intelligent agents and related tasks. 
It is trained to autonomously understand user intent, plan complex instructions, and call one or multiple tools (e.g., web browser, Python interpreter, and the text-to-image model) to complete complex tasks. 
Figure \ref{fig:alltools-arch} presents the overall pipeline of the GLM-4 All Tools system. 
When a user issues a complex request, the model analyzes the task and plan the problem-solving process step by step. 
If it determines that it cannot complete the task independently, it will sequentially call one or multiple external tools, utilizing their intermediate feedback and results to help solve the task.   

Built on the GLM-4's all-tools capabilities, we also developed the GLMs application platform that allows users to create and customize their own agents for specific tasks. 
The GLMs support not only the embedded Python interpreter, web browser, text-to-image model but also user-defined functions, APIs, and external knowledge bases to more effectively address user needs.

\newcommand{\eqwd}{0.1}
\newcommand{\sss}[2]{\shortstack[c]{#1\\{#2}}}
\newcommand{\ddd}[2]{\shortstack[c]{#1\\\tiny{#2}}}

\section{GLM-4 Capabilities}

We examine the capabilities of the GLM-4 model from diverse perspectives, including the base capacity on academic benchmarks, code problem-solving, agent abilities in English, and instruction following, long context for both Chinese and English, as well as alignment in Chinese. 
As mentioned, GLM-4 was pre-trained mostly  in Chinese and English and aligned predominantly to Chinese. 
In this section, we report results primarily for the latest GLM-4 version, i.e., GLM-4 (0520) and GLM-4-Air (0605), as GLM-4 (0520) is slightly better than its original 0116 version across the evaluated benchmarks. 
During evaluation, both GLM-4 and GLM-4-Air are deployed with BFloat16 precision.

For baselines, we present results for GPT-4 (0603), GPT-4 Turbo (1106, 2024-04-09), Claude 2, Claude 3 Opus, and Gemini 1.5 Pro, all of which were extracted from the corresponding technical reports or tested through  their public APIs. 

Overall, GLM-4 gets close to the state-of-the-art models (GPT-4-Turbo, Gemini 1.5 Pro, and Claude 3 Opus) over the standard benchmarks, as well as instruction following, long context, code problem-solving, and agent abilities in English environment. 
For Chinese alignment, it generates strong performance against SOTA models across various domains, such as fundamental language ability, advanced Chinese understanding, professional knowledge, and open-ended question answering.  
In summary, GLM-4 is among the best in terms of Chinese language tasks. 
It also demonstrates comparable performance to GPT-4 and Claude 3 Opus in Chinese math and logic reasoning capabilities though it lags behind GPT-4 Turbo.

\subsection{Evaluation of Academic Benchmarks}

To evaluate the general performance of the base model, we select six commonly-used benchmarks spanning knowledge, math, reasoning, commonsense, and coding:
\begin{itemize}
    \item MMLU~\cite{hendrycks2021measuring}: Multi-choice questions collected from various examinations including mathematics, history, computer science, and more. We present all answers to the model and ask it to choose the letter of the answer. 
    \item GSM8K~\cite{GSM8k:abs-2110-14168}: 8,500 grade school math word problems (1,000 in the test set) that require the model to solve real-life situational problems using mathematical concepts. We use chain-of-thought prompting~\cite{CoT:Wei0SBIXCLZ22} for this benchmark.
        \item MATH: 12,500 challenging competition-level mathematics problems (5,000 in the test set). We use chain-of-thought prompting~\cite{CoT:Wei0SBIXCLZ22} for this benchmark.
    \item BBH~\cite{BBH:SuzgunSSGTCCLCZ23}: A suite of 23 challenging BIG-Bench~\cite{BigBench:abs-2206-04615} tasks. We use chain-of-thought prompting~\cite{CoT:Wei0SBIXCLZ22} for this benchmark.
    \item GPQA~\cite{GPQA:abs-2311-12022}: A graduate-level multi-choice benchmark in biology, chemistry, and physics. 
    \item HumanEval~\cite{Humaneval:abs-2107-03374}: a coding benchmark that measures correctness of synthetic functions with automatic test-case checking. 
\end{itemize}

We compare the performance of GLM-4 with the original GPT-4~\cite{openai2023gpt}. 
The results are shown in \Cref{tab:academic}. 
We can observe that GLM-4 achieves 96.3\% of GPT-4's accuracy on MMLU, and outperforms GPT-4 on other benchmarks. 
Overall, the base capacity of GLM-4 approaches that of GPT-4-Turbo and Claude 3 Opus.

\begin{table}[!ht]
    \centering
    \renewcommand{\arraystretch}{1.5}
    \caption{GLM-4 performance on academic benchmarks.}
    \resizebox{0.8\textwidth}{!}{
    \begin{tabular}{l|cccccc}
    \toprule
    Model & MMLU & GSM8K & MATH & BBH & GPQA & HumanEval\\
    \midrule
    GPT-4 (0314) & 86.4 & 92.0 & 52.9 & 83.1 & 35.7 & 67.0 \\
    GPT-4 Turbo (1106) & 84.7 & 95.7 & 64.3 & 88.3 & 42.5 & 83.7\\
    GPT-4 Turbo (2024-04-09) & 86.7 & 95.6 & 73.4 & 88.2 & 49.3 & 88.2\\
    Claude 3 Opus & 86.8 & 95.0 & 60.1 & 86.8 & 50.4 & 84.9 \\
    Gemini 1.5 Pro & 85.9 & 90.8 & 67.7 & 89.2 & 46.2 & 84.1\\
    \midrule
    GLM-4-9B-Chat & 72.4 & 79.6 & 50.6 & 76.3 & 28.8 & 71.8\\
    GLM-4-Air (0605) & 81.9 & 90.9 & 57.9 & 80.4 & 38.4 & 75.7\\
    GLM-4 (0116) & 81.5 & 87.6 & 47.9 & 82.3 & 35.7 & 72.0 \\
    GLM-4 (0520)  & 83.3 & 93.3 & 61.3 & 84.7 & 39.9 & 78.5\\
    \bottomrule
    \end{tabular}
    }
    \label{tab:academic}
\end{table}

\subsection{Evaluation of Instruction Following}

We assess the proficiency of GLM-4 in following instructions with the recently-introduced IFEval dataset~\cite{zhou2023instruction}.
The dataset comprises 541 prompts derived from 25 distinct instructions that are verifiable through explicit criteria (e.g., \emph{``end your email with: P.S. I do like the cake''} can be verified via string matching).
We adhere to the methodologies outlined by \cite{zhou2023instruction} to calculate prompt-level and instruction-level accuracy in both \emph{strict mode} and \emph{loose mode}.
To further evaluate the model’s performance on following instructions in Chinese, we translate the original prompts into Chinese, omitted instructions that are not applicable in Chinese (such as capitalization), and adjust the scoring scripts to accommodate Chinese data.

\begin{table}[!ht]
    \centering
    \renewcommand{\arraystretch}{1.5}
    \caption{GLM-4 performance on IFEval~\cite{zhou2023instruction}, an LLM instruction following benchmark. `L` stands for `Loose` and `S` stands for `Strict`. `P` stands for `Prompt` and `I` stands for `Instruction`.}
      \resizebox{0.8\textwidth}{!}{
    \begin{tabular}{l|cccc|cccc}
    \toprule
    \multirow{2}{*}{Model} & \multicolumn{4}{c}{English} & \multicolumn{4}{c}{Chinese} \\
    & L-P & S-P & L-I & S-I & L-P & S-P & L-I & S-I \\
    \midrule
    GPT-4 (0613) & 79.5 & 77.1 & 85.5 & 83.7 & 72.4 & 68.9 & 80.0 & 75.7 \\
    GPT-4 Turbo (1106) & 79.1 & 75.4 & 85.1 & 82.4 & 74.3 & 69.1 & 80.8 & 76.5\\
    GPT-4 Turbo (2024-04-09) & 84.5 & 81.2 & 88.7 & 85.9 & 79.3 & 72.6 & 84.2 & 79.1\\
    Claude 2 & 75.0 & 58.0 & 81.7 & 67.7 & 57.1 & 46.5 & 64.9 & 55.1\\
    Claude 3 Opus & 90.6 & 85.5 & 93.7 & 90.0 & 78.3 & 73.3 & 84.3 & 80.4\\
    \midrule
    GLM-4-9B-Chat & 73.0 & 69.0 & 80.3 & 77.2 & 73.0 & 69.0 & 80.3 & 77.2 \\
    GLM-4-Air (0605) & 80.4 & 75.2 & 86.1 & 82.3 & 79.3 & 71.2 & 84.0 & 77.3\\
    GLM-4 (0520) & 83.7 & 79.1 & 88.7 & 85.0 & 79.7 & 71.9 & 84.2 & 78.0\\
    \bottomrule
    \end{tabular}
    }
    \vspace{1em}
    \label{tab:ifeval}
\end{table}

In \emph{loose mode}, GLM-4 matches instruction-level accuracy achieved by GPT-4 Turbo in both English and Chinese.
In \emph{strict mode}, GLM-4 achieves $99.0\%$ and $98.6\%$ of instruction-level accuracy of GPT-4 Turbo (2024-04-09) in English and Chinese, respectively.

\subsection{Evaluation of Alignment}

AlignBench~\cite{liu2023alignbench} provides an automatic LLMs-as-Judge method to benchmark the alignment of LLMs in Chinese context.
It consists 683 queries spanning 8 different categories, and evaluates model responses using a GPT-4 based multidimensional rule-calibrated pointwise reference-based scoring method.
We evaluate on AlignBench-v1.1, which more carefully improves the reference generation quality, especially by complementing human-collected evidences from webpages with urls for knowledge-related questions that takes up 66.5\% of total queries.
On this version, almost all LLMs achieve lower scores than they do in the previous AlignBench.

\begin{table}[!ht]
    \centering
    \renewcommand{\arraystretch}{1.5}
    \caption{GLM-4 performance on AlignBench~\cite{liu2023alignbench}, an LLM benchmark for alignment in Chinese.}
    \resizebox{\textwidth}{!}{
    \begin{tabular}{l|cccccccc|c}
    \toprule
    Model & Math & Logic & Language & Chinese & QA & Writing & Role Play & Professional & Overall \\
    \midrule
    GPT-4 (0613) & 7.54 & 7.17 & 7.82 & 7.02 & 7.39 & 7.67 & 8.20 & 7.29 & 7.46 \\
     GPT-4 Turbo (1106) & 7.85 & 7.66 & 7.90 & 7.22 & 8.24 & 8.53 & 8.46 & 7.95 & 7.90\\
   GPT-4 Turbo (2024-04-09) & 8.32 & 7.67 & 7.60 & 7.57 & 8.37 & 7.75 & 8.18 & 8.59 & 8.00\\
   Claude 2 & 6.39 & 5.85 & 6.75 & 5.72 & 6.68 & 5.87 & 6.86 & 6.56 & 6.26\\
   Claude 3 Opus & 7.27 & 7.11 & 7.94 & 7.71 & 8.21 & 7.61 & 7.73 & 8.02 & 7.53\\
   Gemini 1.5 Pro & 7.07 & 7.77 & 7.31 & 7.22 & 8.55 & 7.83 & 7.79 & 8.52 & 7.47 \\
   \midrule
   GLM-4-9B-Chat & 7.00 & 6.01 & 6.69 & 7.26 & 7.97 & 7.59 & 8.10 & 7.52 & 7.01 \\
   GLM-4-Air (0605) & 7.69 & 6.95 & 7.53 & 8.00 & 7.90 & 8.01 & 8.35 & 8.09 & 7.65\\
   GLM-4 (0116) & 7.20 & 7.20 & 7.60 & 8.19 & 8.45 & 7.88 & 8.05 & 8.56 & 7.66 \\
   GLM-4 (0520) & 7.89 & 7.95 & 8.00 & 7.86 & 8.11 & 8.04 & 8.06 & 8.47 & 8.00\\
   \bottomrule
    \end{tabular}
}
    \label{tab:alignbench}
\end{table}

Results are shown in \Cref{tab:alignbench}.
GLM-4 outperforms GPT-4 Turbo, Claude 3 Opus, and Gemini 1.5 Pro in general, achieves the highest overall score among the baselines.
Especially on Chinese Logic Reasoning and Language Understanding tasks, GLM-4 significantly outperforms all other powerful models.
These results demonstrate its strong grasping of Chinese language and knowledge.

The current performance gap between GLM-4 and GPT-4 Turbo (2024-04-09) mostly lies in the Mathematics dimension.
We have been employing techniques introduced in ChatGLM-Math~\cite{xu2024chatglmmath} such as self-critique to continuously enhance GLM models' math reasoning capabilities.

\subsection{Evaluation of Long Context Handling Abilities}

To assess the performance of GLM-4 on long text tasks, we carry out evaluations on LongBench-Chat~\cite{bai2024longalign}, a benchmark set with context lengths ranging from 10-100k, encompassing a wide range of long text scenarios frequently utilized by users, such as document Q\&A, summarization, and coding. In our quest to provide a more detailed comparison against the performance of GLM-4 in different languages, we also segregate LongBench-Chat according to language. This yields two distinct portions: Chinese and English. We therefore report the results for both segments separately, offering a fine-grained overview of GLM-4's cross-linguistic capabilities.

Regarding the specific evaluation settings, we score the outputs of each model based on GPT-4, adopting a few-shot strategy within LongBench-Chat. Moreover, given our objective to minimize score variations and to reach a more reliable statistical conclusion, we repeated evaluations multiple times. Subsequently, we report the average  from these multiple evaluations in \Cref{tab:longbench} to ensure that the final performance metric reflects a thorough understanding of how GLM-4 behaves under diverse conditions. And the results  clearly suggested that the performance of GLM-4 aligns with that of GPT-4 Turbo and Claude 3 Opus on English prompts, and it outperforms the best of them on Chinese prompts. 

\begin{table}[!ht]
    \centering
    \renewcommand{\arraystretch}{1.5}
    \caption{GLM-4 performance on LongBench-Chat~\cite{bai2023longbench}.}
    \resizebox{0.45\textwidth}{!}{
    \begin{tabular}{l|cc}
    \toprule
       Model  & English & Chinese \\
    \midrule
       GPT-4 Turbo (1106) & 87.2 & 71.4\\
       GPT-4 Turbo (2024-04-09) & 85.0 & 82.1\\
       Claude 2 & 81.3 & 76.2 \\
       Claude 3 Opus & 87.7 & 82.7 \\
       \midrule
       GLM-4-9B-Chat & 76.8 & 79.0\\
       GLM-4-Air (0605) & 82.4 & 81.0\\
       GLM-4 (0520)  & 87.3 & 84.0 \\
       \bottomrule
    \end{tabular}
    }
    \label{tab:longbench}
\end{table}


\subsection{Evaluation of Coding Abilities on Real-world User Prompts}

While HumanEval~\cite{Humaneval:abs-2107-03374} has been widely adopted for evaluating LLMs' code generation, most of its problems are about introductory algorithms.
However, in practice, users ask complicated questions to complete their daily work, whose difficulty is usually far beyond the scope of HumanEval.
Additionally, previous works have reported HumanEval-contaminated training data~\cite{openai2023gpt,li2023textbooks,yang2023rethinking} in their own or other LLMs, making the results on HumanEval relatively less trustful than before.

As a result, beside HumanEval we evaluate GLM-4 on NaturalCodeBench (NCB)~\cite{zhang2024naturalcodebench}, a challenging bilingual coding benchmark derived from real user prompts to mirror the complexity of real-world coding tasks.
As shown in \Cref{tab:ncb}, GLM-4 has a close coding performance to Claude 3 Opus in practical scenarios.
While there is still some gaps to GPT-4 models, considering GLM-4 bilingually balanced nature, there is  quite much potential to improve its performance on NCB via better training strategies and data curation in our following iterations.

\begin{table}[!ht]
    \centering
    \renewcommand{\arraystretch}{1.5}
    \caption{GLM-4 performance on NaturalCodeBench (NCB)~\cite{zhang2024naturalcodebench}, a benchmark with real coding prompts in two programming languages (Python and Java) for English and Chinese.}
    \resizebox{0.8\textwidth}{!}{
    \begin{tabular}{l|cccc|c}
    \toprule
    Model  &  Python (en) & Java (en) & Python (zh) & Java (zh) & Overall\\
    \midrule
    GPT-4 (0613) & 55.7 & 51.1 & 53.4 & 51.1 & 52.8 \\
    GPT-4 Turbo (1106) & 51.9 & 55.0 & 47.3 & 51.9 & 51.5\\
    GPT-4 Turbo (2024-04-09) & 57.5 & 52.3 & 53.1 & 52.3 & 53.8\\
    Claude 2 & 34.4 & 36.6 & 33.6 & 32.8 & 34.4\\
    Claude 3 Opus & 48.9 & 48.9 & 45.0 & 50.4 & 48.3\\
    Gemini 1.5 Pro & 45.0 & 39.7 & 41.5 & 43.1 & 42.3 \\
    \midrule
    GLM-4-9B-Chat & 33.9 & 29.8 & 30.8 & 34.4 & 32.2 \\
    GLM-4-Air (0605) & 40.8 & 39.7 & 43.1 & 39.7 & 40.8\\
    GLM-4 (0520) & 51.6 & 42.8 & 45.4 & 48.9 & 47.1\\
       \bottomrule
    \end{tabular}
    }
    \label{tab:ncb}
\end{table}

\subsection{Evaluation of Function Call}

To evaluate the performance of GLM models on function call, we carry out evaluations on Berkeley Function Call Leaderboard~\cite{berkeley-function-calling-leaderboard}, a benchmark with 2k question-function-answer pairs. The benchmark evaluates model's ability on calling functions in three categories: evaluation by Abstract Syntax Tree (AST), evaluation by executing APIs, and relevance detection. The first category compares the model output functions against function documents and possible answers with AST analysis. The second category checks for response correctness by executing the generated function calls. Relevance detection evaluates the model's capacity on recognizing functions that are not suitable to address the user's question. 
The results are shown in \Cref{tab:fc_result}. We can observe that the function-call capability of  GLM-4 (0520) aligns with that of GPT-4 Turbo (2024-04-09), while GLM-4-9B-Chat significantly outperforms Llama-3-8B-Instruct. Another observation is that the overall accuracy does not improve with model sizes, while GLM-4-9B-Chat can even outperform GLM-4-Air. 
On the other hand, we observe that the performance on execution summary, which evaluates the execution results of real-world APIs, improves smoothly with model size.

\begin{table}[!ht]
    \centering
    \renewcommand{\arraystretch}{1.5}
    \caption{GLM performance on the Berkeley Function Call Leaderboard.}
    \resizebox{0.8\columnwidth}{!}{
    \begin{tabular}{l|ccc|c}
        \toprule
        Model & AST Summary & Exec Summary & Relevance & Overall  \\
        \midrule
        Llama-3-8B-Instruct & 59.25 & 70.01 & 45.83 & 58.88 \\
        GPT-4 Turbo (2024-04-09) & 82.14 & 78.61 & 88.75 & 81.24\\
        GPT-4o (2024-05-13) & 85.23 & 80.37 & 81.25 & 82.94\\
        \midrule
        ChatGLM3-6B & 62.18 & 69.78 & 5.42 & 57.88 \\
        GLM-4-9B-Chat & 80.26 & 84.40 & 87.92 & 81.00 \\
        GLM-4-Air (0605) & 84.34 & 85.93 & 68.33 & 80.94\\
        GLM-4 (0520) & 82.59 & 87.78 & 84.17 & 81.76\\
        \bottomrule
    \end{tabular}
    }
    \label{tab:fc_result}
\end{table}

\subsection{Evaluation of Agent Abilities}

It is widely observed that LLMs are capable to serve as intelligent agents in versatile environments and contexts~\cite{park2023generative,yao2022react}, known as LLMs-as-Agents~\cite{liu2023agentbench}.
As a result, we evaluate GLM-4 together with other comparison LLMs on AgentBench~\cite{liu2023agentbench}, a comprehensive agentic benchmark for text-based LLMs across an array of practical environments, including code-based, game-based, and web-based contexts.
Specifically, we evaluate on 7 out of 8 AgentBench environments except for Digital Card Game, which is too time-consuming to interact with.
Overall scores are calculated using the original per-dataset weights provided in AgentBench~\cite{liu2023agentbench}.

\begin{table}[!ht]
\footnotesize
\renewcommand{\arraystretch}{1.5}
\caption{GLM-4 performance on AgentBench~\cite{liu2023agentbench}.}
\renewcommand\tabcolsep{2pt}
\resizebox{\columnwidth}{!}{
\begin{tabular}{@{}l|ccccccc|c@{}}
\toprule
\multicolumn{1}{c|}{}  & \begin{tabular}[c]{@{}c@{}}Operating\\ System\end{tabular} & DataBase & \begin{tabular}[c]{@{}c@{}}Knowledge\\ Graph\end{tabular} & \begin{tabular}[c]{@{}c@{}}Lateral Thinking\\ Puzzles\end{tabular} & \begin{tabular}[c]{@{}c@{}}House\\ Holding\end{tabular} & \begin{tabular}[c]{@{}c@{}}Web\\ Shopping\end{tabular} & \begin{tabular}[c]{@{}c@{}}Web\\ Browsing\end{tabular} & Overall \\ \midrule
GPT-4 (0613)                 & 42.4                                                       & 32.0     & 58.8                                                      & 16.6                                                               & 78.0                                                    & 61.1                                                   & 29.0                                                   & 3.69    \\
GPT-4 Turbo (1106)       & 40.3                                                       & 52.7     & 54.0                                                      & 17.7                                                               & 70.0                                                    & 52.8                                                   & 30.0                                                   & 3.77    \\
GPT-4 Turbo (2024-04-09) & 41.0                                                       & 46.7     & 53.2                                                      & 19.4                                                               & 72.0                                                    & 55.1                                                   & 19.0                                                   & 3.68    \\
Claude 2               & 18.1                                                       & 27.3     & 41.3                                                      & 8.4                                                                & 54.0                                                    & 61.4                                                   & 0.0                                                    & 2.03    \\
Claude 3 Opus          & 23.6                                                       & 55.0     & 53.4                                                      & 20.0                                                               & 70.0                                                    & 48.5                                                   & 28.0                                                   & 3.62    \\ \midrule
GLM-4-Air (0605)              & 31.9                                                       & 51.0     & 53.8                                                      & 12.3                                                               & 78.0                                                    & 69.2                                                   & 30.0                                                   & 3.58    \\
GLM-4 (0520)           & 36.8                                                       & 52.7     & 51.4                                                      & 15.3                                                               & 82.0                                                    & 68.3                                                   & 29.0                                                   & 3.79    \\ \bottomrule
\end{tabular}
}
\label{tab:agent}
\end{table}

The results are presented in Table~\ref{tab:agent}.
As it shows, GLM-4 models present quite impressive performance on agent tasks, with the GLM-4-Air's comparable and GLM-4's outperforming results to GPT-4 Turbo and Claude 3 Opus.
In terms of specific environments, we find GLM-4 series performed especially well on Database, House-Holding, and Web Shopping tasks, while still demonstrating a gap to GPT-4 series on Operating System, Knowledge Graph, and Lateral Thinking Puzzles.
The gap suggests that there is still room for GLM-4 to improve its performance on code-related agentic tasks and highly interactive language tasks.

\subsection{Evaluation of All Tools}

GLM-4 is further aligned to support intelligent agents and user-configured GLMs functionalities on \url{https://chatglm.cn}, and the resultant model is GLM-4 All Tools. 
As mentioned, GLM-4 All Tools can  complete complex tasks by autonomously understanding user intent, planing step-by-step instructions, and calling multiple tools, including web browser, Python interpreter, and the text-to-image model (e.g., CogView3~\cite{zheng2024cogview3}. 
\Cref{tb:alltools} shows that GLM-4 All Tools (Web) achieved similar performance on Python interpreter for solving math problems, browser for information seeking, compared to ChatGPT-4 (Web), respectively. 

\begin{table}[!ht]
\centering
\renewcommand{\arraystretch}{1.5}
\caption{Performance of GLM-4 All Tools.}
\resizebox{0.7\columnwidth}{!}{
\begin{tabular}{c|c|cc}
\toprule
 & & {GLM-4 All Tools}  & {GPT-4}\\
 & & (Web, 0116) & (Web, 0110)\\
\midrule
\multirow{3}{*}{\shortstack[l]{{Python }\\{Interpreter}   }}  
 & GSM8K    & 91.59  & 92.72   \\
 & MATH     & 63.60  & 65.00   \\
 & Math23K  & 88.50  & 88.40   \\
 \midrule
\multirow{1}{*}{\shortstack[l]{{Browser } }}  
& Information Seeking & 78.08  & 67.12 \\

\bottomrule
\end{tabular}
}
\label{tb:alltools}
\end{table}

\section{Safety and Risks}

We are committed to ensuring that GLM-4 operates as a safe, responsible, and unbiased model. 
In addition to addressing common ethical and fairness concerns, we carefully assess and mitigate potential harms that the model may pose to users in real-world scenarios.

\begin{table*}[!h]
\centering
\renewcommand{\arraystretch}{1.5}
\renewcommand\tabcolsep{2.5pt}
\caption{GLM-4 performance on SafetyBench \cite{zhang2023safetybench}, compared to GPT-4 models and Claude 3 Opus.}
\resizebox{1.0\textwidth}{!}{
\small
\begin{tabularx}{\textwidth}{l|ccccccc|c }
\toprule
  & \begin{tabular}[c]{@{}c@{}}Ethics \&\\ Morality\end{tabular} 
  & \begin{tabular}[c]{@{}c@{}}Illegal\\ Activities\end{tabular} 
  & \begin{tabular}[c]{@{}c@{}}Mental\\ Health\end{tabular} 
  & \begin{tabular}[c]{@{}c@{}}Offens-\\iveness \end{tabular}
  & \begin{tabular}[c]{@{}c@{}}Physical\\ Health\end{tabular} 
  & \begin{tabular}[c]{@{}c@{}}Privacy \&\\ Property\end{tabular} 
  & \begin{tabular}[c]{@{}c@{}}Unfairness\\\& Bias\end{tabular} 
  & Overall \\
\midrule

{GPT-4 (0613)}  & 92.7 & 93.3 & 93.0 & 87.7 & 96.7 & 91.3 & 73.3 & 89.7 \\
{GPT-4 Turbo (1106)} & 91.0 & 92.0 & 93.0 & 86.0 & 92.0 & 88.7 & 74.3 & 88.1 \\
{GPT-4 Turbo (2024-04-09)} & 90.3 & 91.3 & 91.7 & 85.3 & 92.0 & 89.3 & 75.0 & 87.9 \\
{Claude 3 Opus} & 92.7 & 91.7 & 92.7 & 86.3 & 94.7 & 88.7 & 66.0 & 87.5 \\
\midrule
{GLM-4 (0520)}  & 92.3 & 91.3 & 93.3 & 86.3 & 92.3 & 88.6 & 66.0 & 87.2 \\
\bottomrule
\end{tabularx}
}
\label{tb:safety}
\end{table*}

\vpara{Risk Mitigation.}
We carefully cleaned data in the pre-training stage by removing text containing sensitive keywords and web pages from a pre-defined blacklist.
In the alignment phase, we evaluate each training sample for safety and remove any that pose potential risks.
Harmlessness is also an important criteria for preference alignment when comparing multiple model outputs.

We have a red team that constantly challenges the model with tricky questions that tend to cause unsafe answers.
We collect all harmful question-answer pairs from GLM-4 and improve them with human annotations for further model alignment.

\newcommand{\safetyfont}[1]{{\emph{#1}}}

\vpara{Safety Evaluation.}
We evaluate the GLM-4 model on the SafetyBench \cite{zhang2023safetybench}, which assesses each model from 7 dimensions: 
\safetyfont{Ethics and Morality} (unethical behaviors), 
\safetyfont{Illegal Activities} (basic knowledge of law), 
\safetyfont{Mental Health} (adverse impacts on mental health), 
\safetyfont{Offensiveness} (offensive behaviors), 
\safetyfont{Physical Health} (dangerous behaviors that can cause physical harms), 
\safetyfont{Privacy and Property} (privacy breach or property loss), 
\safetyfont{Unfairness and Bias}.
We evaluate different models on the Chinese subset of SafetyBench, which is created by removing highly sensitive questions that tend to be censored, to mitigate interference from different API safety policies.

Table \ref{tb:safety} shows the safety results of GLM-4 and SOTA models.
On most dimensions GLM-4 (0520) shows competitive safety performance, and overall it achieves comparable performance with Claude 3 Opus.
GLM-4 slightly falls behind the GPT-4 family, especially on the Physical Health dimension, which demands robust common sense knowledge about the physical world to avoid potential risks.
More efforts have been put into this direction to develop a more capable and safe GLM model.

\section{Conclusion}

In this report, we introduce the ChatGLM family of large language models from GLM-130B to GLM-4 (All Tools). 
Over the past one and half years, we have made great progress in understanding various perspectives of large language models from our first-hand experiences. 
With the development of each model generation, the team has learned and applied more effective and efficient  strategies for both model pre-training and alignment. 
The recent ChatGLM models---GLM-4 (0116, 0520), GLM-4-Air (0605), and GLM-4 All Tools---demonstrate significant advancements in understanding and executing complex tasks by autonomously employing external tools and functions. 
These GLM-4 models have achieved performance on par with, and in some cases surpassing, state-of-the-art models such as GPT-4 Turbo, Claude 3 Opus, and Gemini 1.5 Pro, particularly in handling tasks relevant to the Chinese language. 
In addition, we are committed to promoting accessibility and safety of LLMs through open releasing of our model weights and techniques developed throughout this journey. 
Our open models, including language, code, and vision models, have attracted over 10 million downloads on Hugging Face in the year 2023 alone. 
Currently, we are working on more capable models with everything we have learned to date. 
In the future, we will continue democratizing cutting-edge LLM technologies through open sourcing, and push the boundary of model capabilities towards the mission of teaching machines to think like humans.

\vpara{Acknowledgement.} 
We would like to thank all the data annotators, infra operating staffs, collaborators, and partners as well as everyone at Zhipu AI and Tsinghua University not explicitly mentioned in the report who have provided support, feedback, and contributed to ChatGLM. 
We would also like to thank Yuxuan Zhang and Wei Jia from Zhipu AI as well as the teams at Hugging Face, ModelScope, WiseModel, and others for their help on the open-sourcing efforts of the GLM family of models.

\bibliographystyle{abbrv}
\bibliography{ref}

\end{document}